\documentclass[runningheads]{llncs}
\usepackage{graphicx}
\usepackage{multicol}
\usepackage{hyperref}
\usepackage{multirow}
\usepackage{amsmath,bm}
\usepackage{bbding}
\usepackage{color}
\usepackage{hyperref}
\hypersetup{
    colorlinks=true,
    linkcolor=blue,
    filecolor=blue,      
    urlcolor=magenta,
    citecolor=blue,
}
\begin{document}
\title{HDTR-Net: A Real-Time High-Definition Teeth Restoration Network for Arbitrary Talking Face Generation Methods}
\author{Yongyuan Li\inst{1,2} \and
Xiuyuan Qin\inst{3} \and
Chao Liang\inst{4} \and
Mingqiang Wei\inst{1,2}\textsuperscript{\Envelope}}

\institute{Nanjing University of Aeronautics and Astronautics, Nanjing, China \and
Shenzhen Research Institute, Nanjing University of Aeronautics and Astronautics, Shenzhen, China \and
Soochow University, Suzhou, China
\and
Nanjing University of Science and Technology, Nanjing, China \\
\textsuperscript{\Envelope}mqwei@nuaa.edu.cn
}
\maketitle              
\begin{figure*}[ht]
\centering
\includegraphics[width=0.9\textwidth]{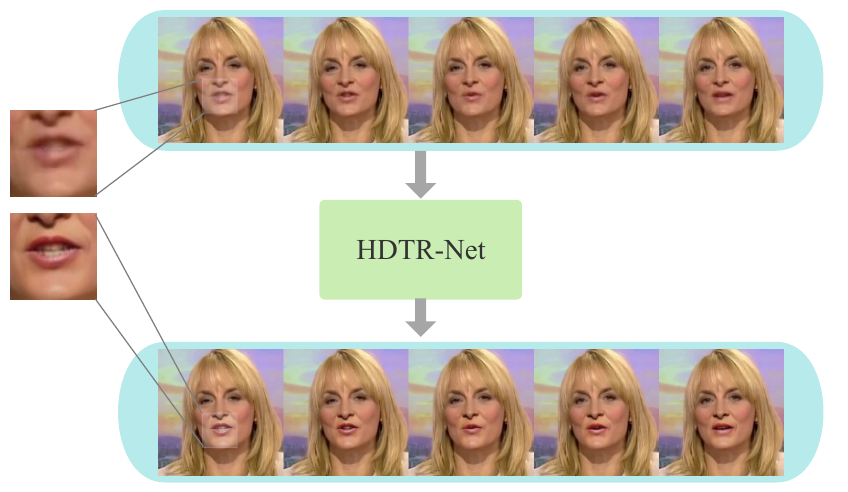}
\caption{Our method effectively enhances the clarity of teeth and surroundings generated by arbitrary talking face videos.}
\label{teaser}
\end{figure*}
\vspace{-0.8cm} %调整图片与上文的垂直距离

\begin{abstract}
% The abstract should briefly summarize the contents of the paper in
% 15--250 words.
Talking Face Generation (TFG) aims to reconstruct facial movements to achieve high natural lip movements from audio and facial features that are under potential connections. 
% Its main objective is achieving high natural lip synchronization. 
Existing TFG methods have made significant advancements to produce natural and realistic images. However, most work rarely takes visual quality into consideration. 
It is challenging to ensure lip synchronization while avoiding visual quality degradation in cross-modal generation methods. To address this issue, we propose a universal High-Definition Teeth Restoration Network, dubbed HDTR-Net, for arbitrary TFG methods. HDTR-Net can enhance teeth regions at an extremely fast speed while maintaining synchronization, and temporal consistency. In particular, we propose a Fine-Grained Feature Fusion (FGFF) module to effectively capture fine texture feature information around teeth and surrounding regions, and use these features to fine-grain the feature map to enhance the clarity of teeth. Extensive experiments show that our method can be adapted to arbitrary TFG methods without suffering from lip synchronization and frame coherence. Another advantage of HDTR-Net is its real-time generation ability. 
Also under the condition of high-definition restoration of talking face video synthesis, its inference speed is $300\%$ faster than the current state-of-the-art face restoration based on super-resolution. Our code and trained models are released at
\url{https://github.com/yylgoodlucky/HDTR}.

\end{abstract}

\section{Introduction}

Talking Face Generation (TFG) plays an important role in the audio-visual field~\cite{TFG-Survey}, as it enables the integration of visual and auditory information to enhance the understanding and perception of information for humans.

For TFG, a clear and realistic mouth in the generated image could provide a richer audio-visual experience and thus help the user to understand the semantics better. 
Traditional TFG methods warp the source image with the help of prior knowledge (e.g., audio), and many of them result in inaccurate output. 
Thanks to the successful application of deep neural networks, especially Generative Adversarial Networks (GAN) \cite{LeeCH22} and Convolution Neural Networks (CNN), TFG has made significant progress. Some efforts \cite{abs-2203-03984,ZhuHLZH20,PrajwalMNJ20,Zhou000W19,VougioukasPP19,SongZLWQ19,iplap} try to produce natural and realistic talking faces by extracting driving features from speech signals and then integrate them into face animation. 
Existing TFG methods focus on producing natural realistic and high synchronization mouth shapes. However, it is still challenging to enhance the teeth clarity while ensuring the natural synchronization of mouth shapes.

Analyzing from the data perspective, the low-resolution of images in existing TFG datasets limits the ability of cutting-edge models to generate high-resolution mouth shapes. Analyzing from the network perspective, Obamanet et al. \cite{SuwajanakornSK17} propose a Teeth Proxy to obtain the high-frequency components of the teeth from the candidate frames to improve the clarity of the upper and lower teeth. However, these methods require a rigid selection of the candidate frames and do not optimize the clarity of the surrounding areas of the teeth. 
Some other efforts \cite{PrajwalMNJ20,ZhouSWL0021} use additional reference frames to compensate mouth shapes and motion to guide the network for accurate modeling of head posture and mouth synchronization, which produce excellent lip synchronization, but the model still falls short in terms of clarity in predicting teeth and their surrounding regions. We argue that prior knowledge is insufficient to provide and restore fine-grained features about the teeth and their surrounding regions.

Most recent efforts \cite{WangLZS21,ChenT0S018,LiLRZ0Z20,WangXDS21} recover high-frequency details from blurred and degraded face images while maintaining the original face features, but it still has gaps in terms of frame coherence and speed. In face restoration, the process is applied to each frame individually, which can result in noticeable pixel discontinuities in the synthesized video. This occurs because face restoration primarily focuses on recovering high-frequency information within each frame, without considering the continuity between frames. As a result, there may be visible inconsistencies or discrepancies in the appearance of consecutive frames, leading to a lack of smoothness or coherence in the synthesized video.

% Face restoration is applied to each frame, and there will be an obvious discontinuity of pixels in the synthesized video, which is because in face restoration only high-frequency information recovery of the image, and the before-and-after continuity of frames is not considered. 
In addition, processing each frame takes a lot of time even on low-resolution images, which limits the practical applicability of face restoration in real-time or time-sensitive scenarios. 

We propose a universal real-time High-Definition Teeth Restoration Network (HDTR-Net). HDTR-Net can be applied to arbitrary Talking Face Generation methods for generating quality results (see Fig. \ref{teaser}). Two components are involved in the proposed method: a Fine-Grained Feature Fusion (FGFF) module and a Decoder module. 
We deliberately design the FGFF module to merge features responsible for extracting the image texture details.  
The branch below the FGFF module is utilized to leverage the reference image as guidance, enabling the model to effectively restore high-frequency details. 
In addition, it can be used as a prompt for pixel continuity in the inference stage. The Decoder module restores the high-frequency details from the extracted fine-grained feature.
The main contributions of our work are three-fold:
\begin{itemize}
    \item  We propose a High-Definition Teeth Restoration Network, dubbed HDTR-Net, to enhance the clarity of teeth regions while maintaining texture details. HDTR-Net can be applied to arbitrary talking face generation.
    \item  We propose a Fine-Grained Feature Fusion module, which has the ability to extract fine-grained features effectively.
    \item  HDTR-Net exhibits exceptional speed in terms of repairing capability. For example, our inference speed is $300\%$ faster than the super-resolution based image restoration methods.
\end{itemize}

\section{Related Work}
% In this section, we will briefly summarize relevant approaches for Talking Face Generation and Face Restoration.

\subsection{Talking Face Generation}
Talking Face Generation aims to synthesize a sequence of talking face frames according to a sequence of driving audio or text, which is a multi-modal learning method for mapping acoustic features to real facial motions. In recent years, \cite{SuwajanakornSK17,obamanet} have learned implicit mapping functions from audio to corresponding landmarks of the mouth to implement talking face synthesis, but their work is only specific to a particular speaker and audio. It is not sufficient to extend to arbitrary speakers and audio, and the synthesized video has low-quality clarity. \cite{JamaludinCZ19} proposes a deblurring module, which uses the idea of early super-resolution \cite{KimLL16a} to transfer the facial feature map from the input image to the generated output using a skip connection to avoid producing a blurred image. Subsequently, \cite{VougioukasPP19,VougioukasPP20,ZhuHLZH20,PrajwalMNJ20} train on the large-scale public dataset (e.g., LRS2 \cite{AfourasCSVZ22}, LRW \cite{ChungSVZ17}) in an end-to-end manner and produce effective results, being able to generalize to arbitrary speakers and audio, but the mouth region is  still of low resolution in the synthesized video, resulting in a worse visual experience for the viewer. \cite{EskimezMXD18,abs-1905-03820,abs-2004-12992} firstly allow the speech to predict the face landmarks, and then recover the real face image. Predicting face landmarks couple speech features and mouth motion features obtains better lip synchronization, but the two-stage training method inherently loses information. In order to synthesize more natural and realistic head motions, the reference sequence images are fed into the network to reduce the head pose and mouth motions \cite{ZhouSWL0021,abs-2303-17480,PrajwalMNJ20,Zhou000W19}. Respectively, \cite{ZhouSWL0021} and \cite{Zhou000W19} decoupled the visual representation space using contrastive learning \cite{NagraniCAZ20} and generative adversarial methods \cite{LeeCH22}, which produce significant improvements in lip synchronization. To obtain high-fidelity talking face videos, \cite{abs-2303-03988} proposes a repair network with an adaptive affine transformation module \cite{Zhang022} to achieve clearly synthetic videos with multi-stage training, 
but the inference speed is slow and the fidelity of the mouth shape depends on the clarity of the input image.

\subsection{Face Restoration}
Based on the general face hallucination \cite{CaoLSLL17,HuangHST17,XuSPZP017,YuFHP18,ChenT0S018}, most methods utilize a geometric prior and a reference prior together to improve performance. However, geometric priors are estimated from low-quality input images, and such geometric priors cannot provide detailed information about the image. Reference prior \cite{LiLYZLY18,LiLRZ0Z20,DoganGT19} relies on images of the same identity, and its ability to recover high-frequency details is reduced for particular image feature with rich appearance. In particular, \cite{KolouriR15} estimates face landmarks before restoring a face and estimating facial pose, while for quite minor facial accurate estimation is difficult. \cite{ZhuLLT16} proposes a unified framework for multi-resolution and dense correspondence field estimation of faces to recover texture details. Recently, deep learning-based models have advanced significantly in image processing tasks and are at present driving the state-of-the-art in face restoration. \cite{LedigTHCCAATTWS17,SonderbyCTSH17} have better reconstruction performance with a generative approach. \cite{YuP16} uses the discriminative generative network to ultra-resolve images by aligning miniature low-resolution face images. \cite{ZhouFCJY15} uses convolution to extract features from blurred images to reconstruct high-definition face images, but their restored faces generate unfaithful outcomes. These methods are essentially based on restoration for single images, with the benefit of being able to recover a scaled-up, high-resolution image, however, the temporal coherence present in the video is not taken into account.

\begin{figure*}[ht]
\centering
\includegraphics[width=1\textwidth]{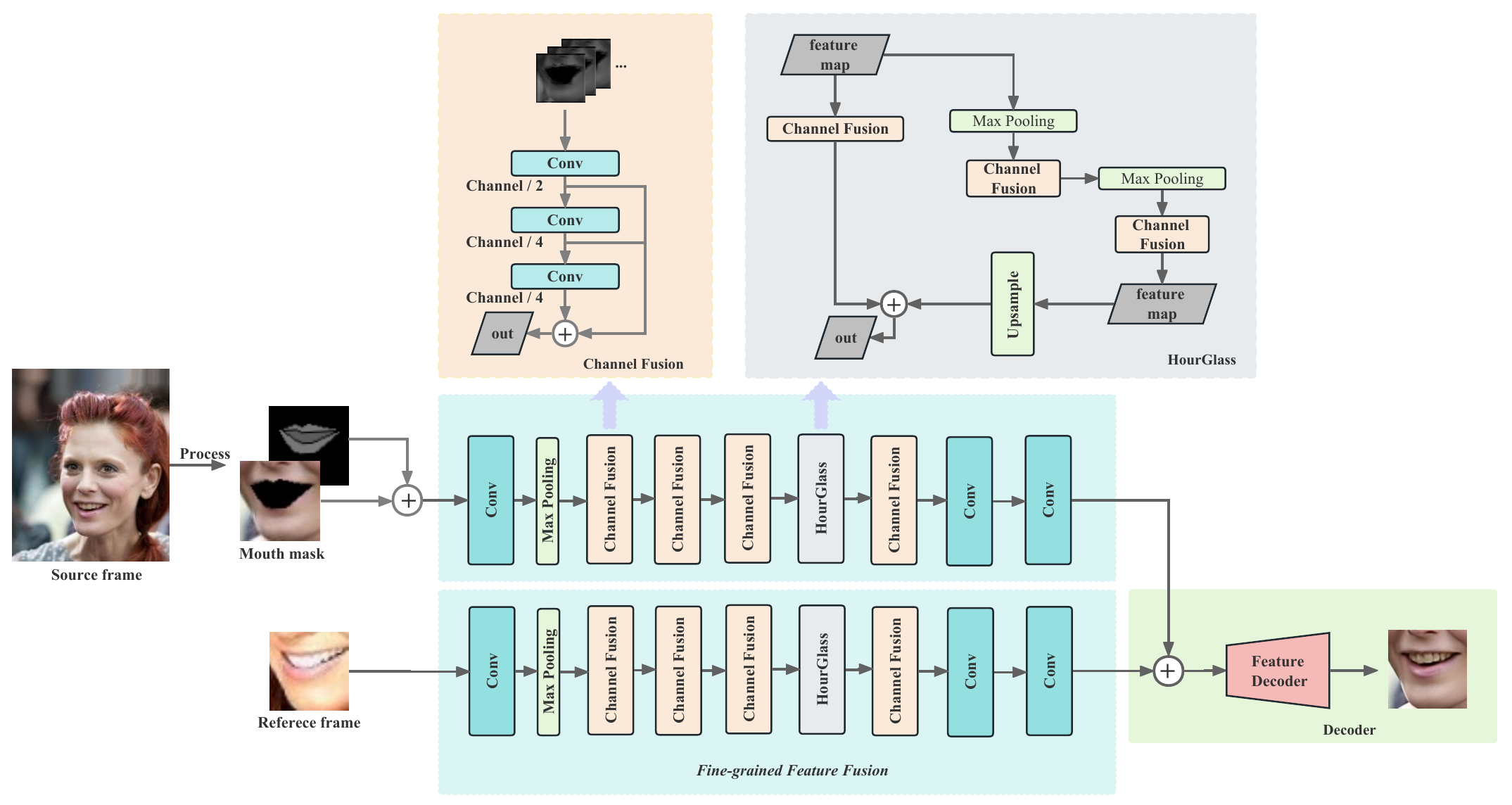}
\vspace{-0.5cm}
\caption{Pipeline of our HDTR-Net. HDTR-Net consists of two parallel Fine-Grained Feature Fusion (FGFF) modules in the blue rectangles and a Decoder module in the green rectangle. Two important components are Channel Fusion (CF) in the orange rectangular and HourGlass in the grey rectangular included in FGFF. FGFF extracts fine-grained texture features from the processed source image and the reference image, then concatenates the feature map to the Decoder for the restoration of teeth regions.}
\label{fig1:HDTR}
\end{figure*}

\section{Method}
We propose a novel real-time teeth restoration network called HDTR-Net that can be adaptive to arbitrary TFG methods. The structural details of HDTR-Net are shown in Fig. \ref{fig1:HDTR}. We first review the structure of the model, which contains two parallel Fine-Grained Feature Fusion (FGFF) modules and a Decoding module. Channel Fusion (CF) and HourGlass are included in FGFF, which focus on extracting fine-grained features.

\subsection{Fine-grained Feature Fusion}
The blue rectangle in Fig. \ref{fig1:HDTR} illustrates the structure of Fine-Grained Feature Fusion (FGFF). Given one source image $I_s \in R^{3 \times H \times W}$, after processing the source image for producing the mouth mask image $I_m \in R^{3 \times H \times W}$ and mouth contour image $I_c \in R^{3 \times H \times W}$, $I_m$ and $I_c$ are fed into the Fine-Grained Feature Fusion (FGFF) module for generating the feature map as:
\begin{equation}
\boldsymbol{f_m}=\mathcal{F}(\boldsymbol{(I_m \oplus I_c)}; {\Theta}_m)
\end{equation}
% \noindent
where $\oplus$ is channel concatenation, after processing the source image, $I_m$ and $I_c$ are concatenated into FGFF, which FGFF module $\mathcal{F}(\cdot ; \Theta_m)$ with a set of parameters $\Theta_m$ transforms $I_m$ and $I_c$ into another feature map $\boldsymbol{f_m}$.

FGFF contains two important components  which are the orange rectangular Channel Fusion (CF) and the grey rectangular HourGlass \cite{hourglass} in Fig. \ref{fig1:HDTR}. CF takes the input of the feature map and passes through three convolutional layers, with the first convolutional layer output a feature map channel that is one-half of the original channel. The last two convolutional layers output a feature map channel that is one-fourth of the original channel. Finally, the output of each convolutional layer is concatenated at the channel dimension, so that the output of CF remains channel invariant but merges with the feature after multi-layer convolutional. Similarly, HourGlass accepts the input of the feature map, embedding the max pooling on the basis of CF. After two layers of max pooling and CF, the features with larger weights are saved and further merged. Finally, the output feature map after upsampling and the CF output feature map are concatenated at the channel dimension. 

Reference image is fed into the branches below FGFF as:
\begin{equation}
\boldsymbol{f_r}=\mathcal{F}(\boldsymbol{I_r}; \Theta_r)
\end{equation}
% \noindent
where $\boldsymbol{I_r}$ is the reference image, FGFF $\mathcal{F}(\cdot ; \Theta_r)$ with a set of parameters $\Theta_r$ transforms $I_r$ into another feature map $\boldsymbol{f_r}$.

\subsection{Decoder}
Decoder focuses on repairing the mouth mask images utilizing two FGFF module results as:
\begin{equation}
\boldsymbol{I_o}=\mathcal{D}((\boldsymbol{f_m} \oplus \boldsymbol{f_r}); \Theta_d)
\end{equation}
where $\oplus$ is channel concatenation, decoder $\mathcal{D}(\cdot ; \Theta_d)$ with a set of parameters $\Theta_d$ transforms $\boldsymbol{f_m}$ and $\boldsymbol{f_r}$ into the output $\boldsymbol{I_o}$.

\subsection{Loss Function}
In the training stage, we use three kinds of loss functions to train HDTR-Net, containing reconstruction loss, perception loss \cite{perceptionloss}, and GAN loss \cite{LeeCH22}.

\textbf{GAN loss}. 
Frame discriminator predicts the probability whether the generated frame is comparable to ground truth, resulting in the loss as:
\begin{equation}
\mathcal{L}_{G A N}=\mathcal{L}_D+\mathcal{L}_G
\end{equation}

\noindent
where
\begin{equation}
\mathcal{L}_D=\frac{1}{2} E\left(D\left(I_g\right)-1\right)^2+\frac{1}{2} E\left(D\left(I_o\right)-0\right)^2
\label{eq1}
\end{equation}
where G represents HDTR-Net and D denotes the discriminator, $\boldsymbol{I_g}$ represents the ground truth image, and $\boldsymbol{I_o}$ represents the teeth restoration image.
% (\ref{eq1})

\textbf{Reconstruction Loss}.
To ensure the coherence of image color and mouth shape, we use L1 loss and L2 loss to reconstruct the mouth region as:
\begin{equation}
\mathcal{L}_{rec}(I_g, I_o)=\left\|I_g-I_o\right\|_1 + \left\|I_g-I_o\right\|_1^2
\end{equation}

\textbf{Perception Loss}.
In order to make the generated image have a more natural appearance, we use perception loss to capture the high-level feature differences between the generated image and the ground truth. We calculate the perception loss in $\boldsymbol{I_g}$ and $\boldsymbol{I_o}$ by pre-training the VGG network \cite{vggnet}, and the perception loss is formulated as:
\begin{equation}
\mathcal{L}_{\text {perc }}\left(I_g, I_o\right)=\left\|\phi\left(I_g\right)-\phi\left(I_o\right)\right\|_2^2
\end{equation}
% \noindent
where $\phi$ is a feature extraction network.

% \noindent
The overall loss is formulated as:
\begin{equation}
\mathcal{L}=\lambda_{1}\mathcal{L}_{GAN}+\lambda_{2}\mathcal{L}_{perc}+\lambda_{3}\mathcal{L}_{rec}
\end{equation}

\begin{figure*}[ht]
\centering
\includegraphics[width=0.8\textwidth]{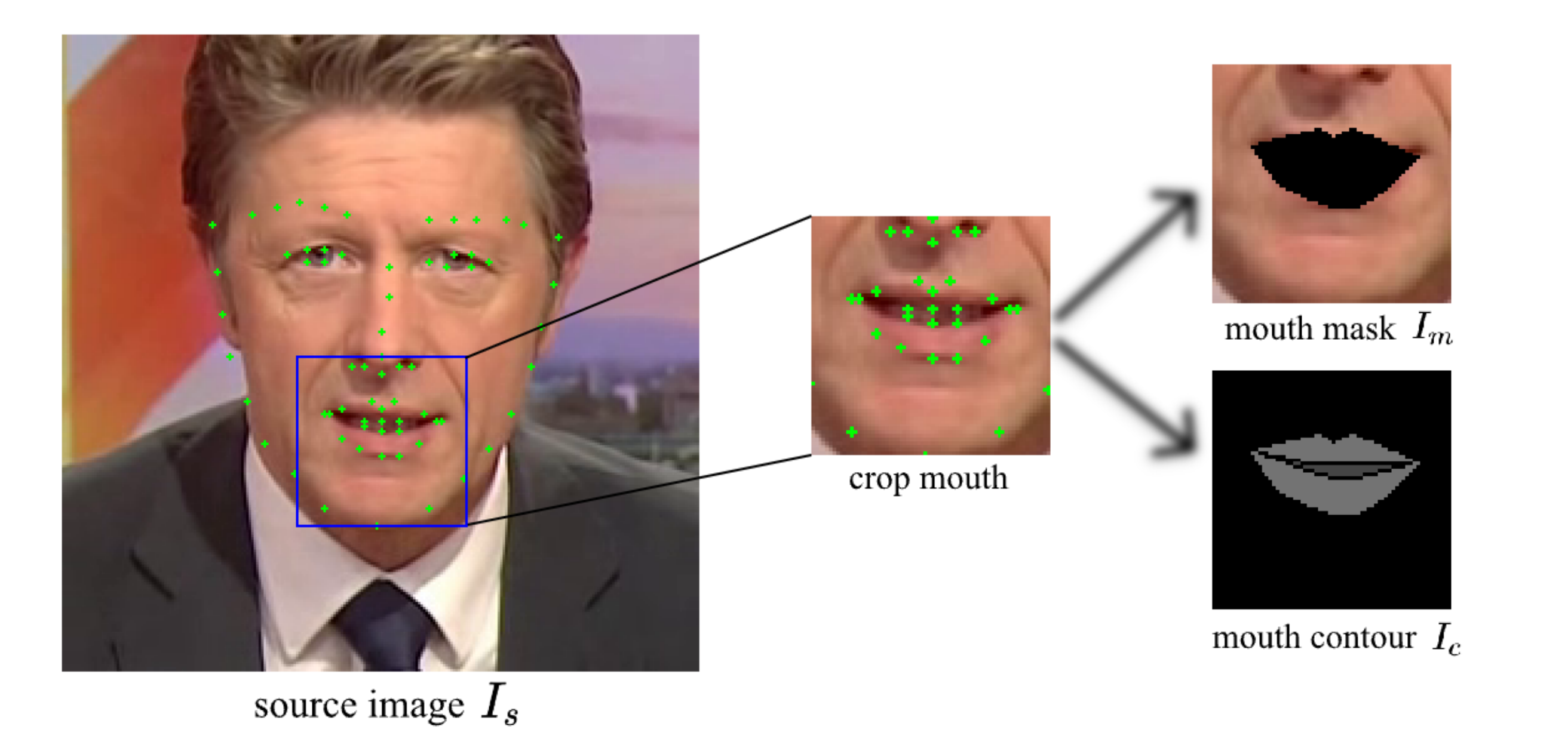}
\caption{Details of data processing. Firstly, face landmarks are extracted from source image $I_s$, then based on the four key points located at the left corner of the mouth, the right corner of the mouth, the tip of the nose, and the jaw, we can determine the boundaries of the mouth region. Using these boundary points, we crop the mouth region from the source image. After processing the cropped mouth region, mouth masks $I_m$ and mouth contour $I_c$ are obained.}
\label{data_process}
\end{figure*}

\section{Experiment}
In this section, we first detail datasets and metrics, comparison methods, and implementation details in our experiment. Then, we show the teeth restoration results of our method. Next, we carry out qualitative and quantitative comparisons with other state-of-the-art works. Finally, we conduct ablation studies. 

\begin{figure*}[ht]
\centering
\includegraphics[width=1\textwidth]{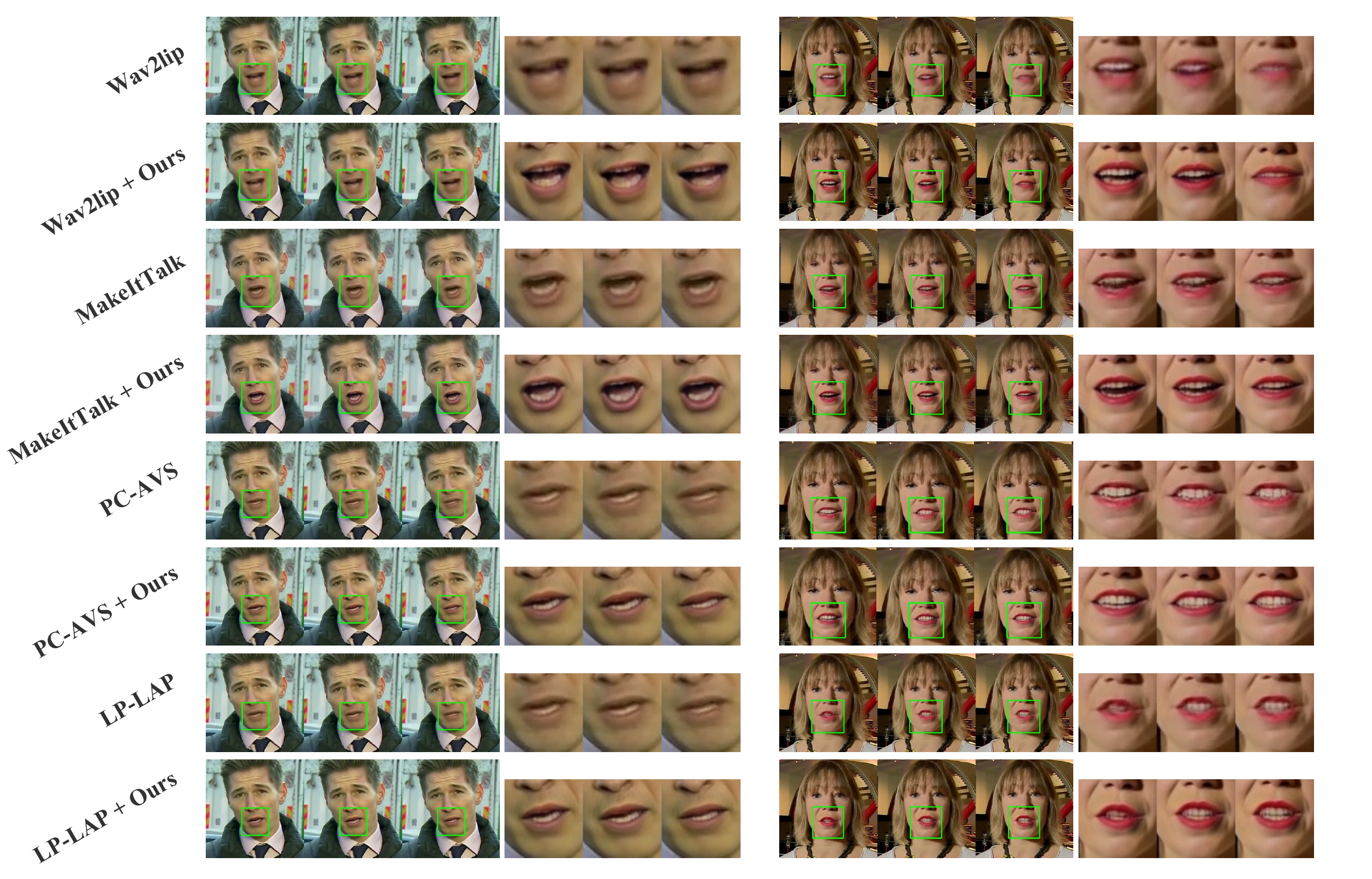}
\caption{Our method is adaptable to arbitrary Talking Face Generation methods, and restores the teeth region to obtain high-definition teeth.}
\label{exp1}
\end{figure*}

\subsection{Experimental Settings}

\subsubsection{Dataset and Metrics}
We train the proposed method in the train set with high definition processed dataset LRS2, and evaluate with other competitive approaches in the test sets of two prevalent benchmark datasets: LRS2 \cite{AfourasCSVZ22} and LRW \cite{ChungSVZ17}. LRS2 contains more than 1000 speakers, with nearly 150,000 instances of words captured and 63,000 different words due to unlimited sentences at the time of capture. LRW selects the 500 most frequent words and clips the speaker's speech, resulting in over 1000 speakers and over 55,000,000 instances of speech.

During the testing phase of our method, there is no ground truth available for direct comparison. To assess the effectiveness of our method, we employ eight image sharpness evaluation metrics: Brenner, Laplacian, SMD, SMD2, Variance, Energy, Vollath, and Entropy. These metrics are chosen to measure the quality of the restored images, aiming to align with human subjective perception. In addition, we measure the time consumption to repair the images.

\subsubsection{Comparison Methods}
We compare our method with super-resolution based face restoration methods, including ESRGAN \cite{esrgan}, GFPGAN \cite{WangLZS21}, and Real-ESRGAN \cite{WangXDS21}. Real-ESRGAN is an enhanced version of the ESRGAN method. These methods input images of different resolutions into the generator to output 1$\times$ and multiples of high-resolution enlarged images, but our method's input and output are all 96$\times$96 resolution. To ensure a fair comparison, we maintain the same input and output resolution for all the comparison experiments, the same as our method.

\subsubsection{Implementation Details}
Data processing is shown in Fig. \ref{data_process}. Given source image $I_s \in R^{3 \times H \times W}$, we begin by extracting face landmarks and cropping the mouth region using four key points: the left corner and the right corner of the mouth, the tip of the nose, and the jaw. Then we obtain a rectangular region containing the mouth and surroundings. Using the extracted face landmarks, we construct a mouth mask $I_m \in R^{3 \times H \times W}$ and mouth contour $I_c \in R^{3 \times H \times W}$ using the key points of the lips. To ensure consistency, $I_m \in R^{3 \times H \times W}$ and $I_c \in R^{3 \times H \times W}$ are aligned to a resolution of $96\times96$  pixels, focusing on the mouth region. These aligned frames are then input into HDTR-Net.

During training, HDTR-Net takes three inputs: one masked mouth image $I_m \in R^{3 \times 96 \times 96}$, one mouth contour image $I_c \in R^{3 \times 96 \times 96}$, and one randomly selected image $I_f \in R^{3 \times 96 \times 96}$ from the training dataset as a reference frame. We use Adam optimizer \cite{adam} with a default setting to optimize HDTR-Net. The learning rate is set to 0.0001. The batch size is set to 12 on one A100 GPU.

\begin{table*}[ht]
  \centering
  \caption{Quantitative comparisons with the state-of-the-art methods on image quality.}
   \renewcommand{\arraystretch}{1.0}
  \resizebox{1.0\linewidth}{!}{
    \begin{tabular}{cccccccccc}
    \hline
    \\
    \multicolumn{1}{c}{Metrics} & \multicolumn{1}{c}{Time $\downarrow$} & \multicolumn{1}{c}{Brenner $\uparrow$} & \multicolumn{1}{c}{Laplacian $\uparrow$} & \multicolumn{1}{c}{SMD $\uparrow$} & \multicolumn{1}{c}{SMD2 $\uparrow$} & \multicolumn{1}{c}{Variance $\uparrow$} & \multicolumn{1}{c}{Energy $\uparrow$} & \multicolumn{1}{c}{Vollath $\uparrow$} & \multicolumn{1}{c}{Entropy $\uparrow$} \\
    \\
    
    % \hline
    % \multicolumn{1}{c}{\multirow{2}[2]{*}{source}} & \multirow{2}[2]{*}{} & \multirow{2}[2]{*}{1859084} & \multirow{2}[2]{*}{92.4198} & \multirow{2}[2]{*}{55275} & \multirow{2}[2]{*}{181290} & \multirow{2}[2]{*}{4426551} & \multirow{2}[2]{*}{713562} & \multirow{2}[2]{*}{3460995} & \multirow{2}[2]{*}{4.5727} \\
    %       &       &       &       &       &       &       &       &       &  \\
    \hline
    \multicolumn{1}{c}{\multirow{2}[2]{*}{ESRGAN}} & \multirow{2}[2]{*}{0.0597} & \multirow{2}[2]{*}{2003566} & \multirow{2}[2]{*}{81.38} & \multirow{2}[2]{*}{\textbf{163871}} & \multirow{2}[2]{*}{193160} & \multirow{2}[2]{*}{33401059} & \multirow{2}[2]{*}{725269} & \multirow{2}[2]{*}{31201016} & \multirow{2}[2]{*}{4.56} \\
      &       &       &       &       &       &       &       &       &  \\
    \multicolumn{1}{c}{\multirow{2}[2]{*}{GFPGAN}} & \multirow{2}[2]{*}{1.2332} & \multirow{2}[2]{*}{2302120} & \multirow{2}[2]{*}{140.18} & \multirow{2}[2]{*}{72312} & \multirow{2}[2]{*}{240881} & \multirow{2}[2]{*}{6318184} & \multirow{2}[2]{*}{983560} & \multirow{2}[2]{*}{5108233} & \multirow{2}[2]{*}{4.56} \\
          &       &       &       &       &       &       &       &       &  \\
    % \hline
    \multicolumn{1}{c}{\multirow{2}[2]{*}{Real-ESRGAN}} & \multirow{2}[2]{*}{0.0622} & \multirow{2}[2]{*}{2029371} & \multirow{2}[2]{*}{75.15} & \multirow{2}[2]{*}{66755} & \multirow{2}[2]{*}{199590} & \multirow{2}[2]{*}{6190525} & \multirow{2}[2]{*}{772949} & \multirow{2}[2]{*}{5069267} & \multirow{2}[2]{*}{4.55} \\
          &       &       &       &       &       &       &       &       &  \\
    % \hline

    \hline
    \multicolumn{1}{c}{\multirow{2}[2]{*}{\textbf{Ours}}} & \multirow{2}[2]{*}{\textbf{0.0187}} & \multirow{2}[2]{*}{\textbf{3121676}} & \multirow{2}[2]{*}{\textbf{102.28}} & \multirow{2}[2]{*}{85607} & \multirow{2}[2]{*}{\textbf{331483}} & \multirow{2}[2]{*}{\textbf{8990130}} & \multirow{2}[2]{*}{\textbf{1264451}} & \multirow{2}[2]{*}{\textbf{7667062}} & \multirow{2}[2]{*}{\textbf{4.75}} \\
          &       &       &       &       &       &       &       &       &  \\
    \hline
    \end{tabular}
    }%
  \label{tab1}%
\end{table*}%

\begin{figure*}[ht]
\centering
\includegraphics[width=0.8\textwidth]{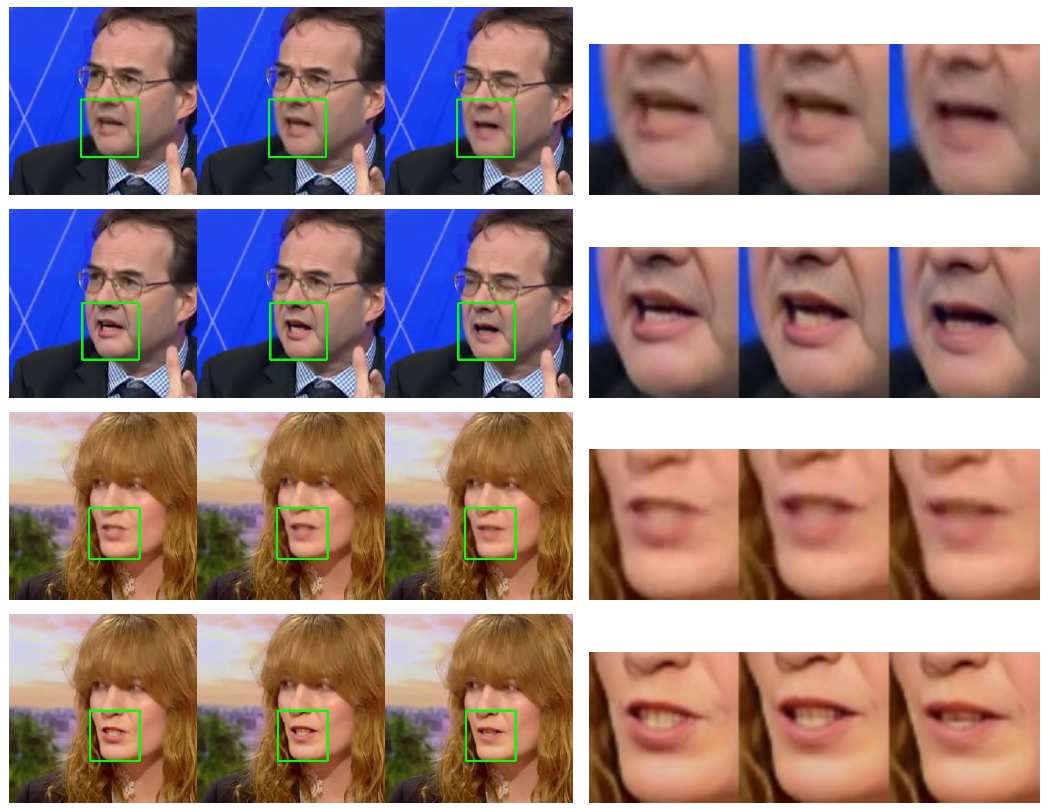}
\caption{Visualization results in side-faced speaker. The first and third rows are the original talking face-generated frames, and the second and fourth rows show the results of the dental restoration applied with our method. It can be clearly seen that our model produces both sharpness and color teeth details in a side-faced speaker, demonstrating the effectiveness and robustness of our method.}
\label{exp_celian}
\end{figure*}

\begin{figure*}[ht]
\centering
\includegraphics[width=0.9\textwidth]{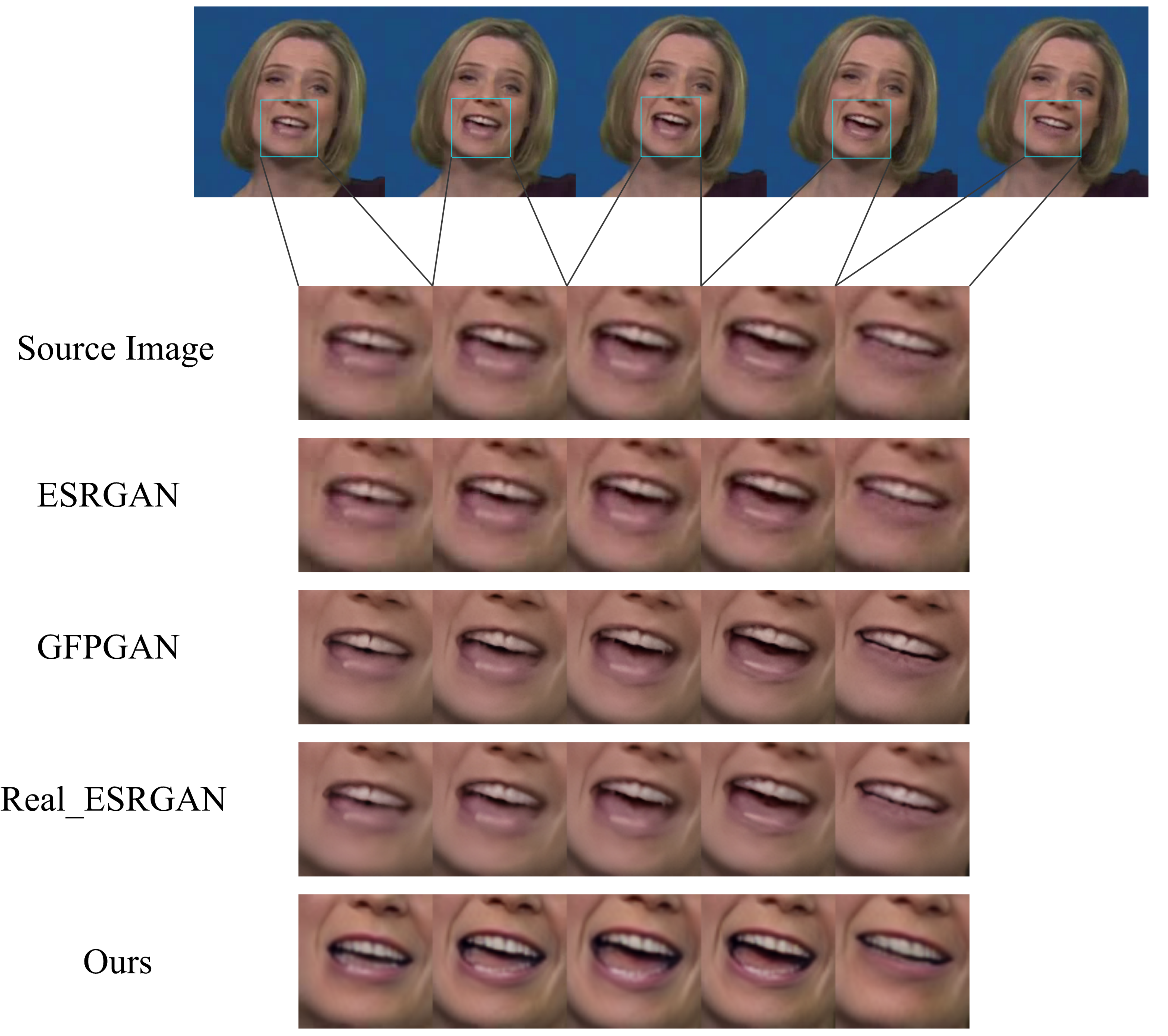}
\caption{Compared to face restoration methods based on super-resolution, the visualization results indicate that our method outperforms significantly in teeth texture detail and color parts.}
\label{exp_hq}
\end{figure*}

\begin{table*}[ht]
  \centering
  \caption{Quantitative results of ablation study.}
   \renewcommand{\arraystretch}{1}
  \resizebox{1.0\linewidth}{!}{
    \begin{tabular}{ccccccccc}
    \hline
       \multicolumn{1}{c}{} &\multicolumn{1}{c}{}&\multicolumn{1}{c}{}&\multicolumn{1}{c}{}&\multicolumn{1}{c}{}&\multicolumn{1}{c}{}&\multicolumn{1}{c}{}&\multicolumn{1}{c}{}&\multicolumn{1}{c}{}
    \\
    
    \multicolumn{1}{c}{Metrics} & \multicolumn{1}{c}{Brenner $\uparrow$} & \multicolumn{1}{c}{Laplacian $\uparrow$} & \multicolumn{1}{c}{SMD $\uparrow$} & \multicolumn{1}{c}{SMD2 $\uparrow$} & \multicolumn{1}{c}{Variance $\uparrow$} & \multicolumn{1}{c}{Energy $\uparrow$} & \multicolumn{1}{c}{Vollath $\uparrow$} & \multicolumn{1}{c}{Entropy $\uparrow$} 
    \\
       \multicolumn{1}{c}{} &\multicolumn{1}{c}{}&\multicolumn{1}{c}{}&\multicolumn{1}{c}{}&\multicolumn{1}{c}{}&\multicolumn{1}{c}{}&\multicolumn{1}{c}{}&\multicolumn{1}{c}{}&\multicolumn{1}{c}{}
    \\
    \hline
    \multicolumn{1}{c}{\multirow{2}[2]{*}{w/o CF}} & \multirow{2}[2]{*}{2101920} & \multirow{2}[2]{*}{68.18} & \multirow{2}[2]{*}{68294} & \multirow{2}[2]{*}{253783} & \multirow{2}[2]{*}{6392508} & \multirow{2}[2]{*}{923189} & \multirow{2}[2]{*}{6912671} & \multirow{2}[2]{*}{4.21} \\
         \multicolumn{1}{c}{} &\multicolumn{1}{c}{}&\multicolumn{1}{c}{}&\multicolumn{1}{c}{}&\multicolumn{1}{c}{}&\multicolumn{1}{c}{}&\multicolumn{1}{c}{}&\multicolumn{1}{c}{}&\multicolumn{1}{c}{}
          \\
    % \hline
    \multicolumn{1}{c}{\multirow{2}[2]{*}{w/o ref FGFF}} & \multirow{2}[2]{*}{2803462} & \multirow{2}[2]{*}{86.12} & \multirow{2}[2]{*}{74029} & \multirow{2}[2]{*}{248926} & \multirow{2}[2]{*}{7183411} & \multirow{2}[2]{*}{871762} & \multirow{2}[2]{*}{6727348} & \multirow{2}[2]{*}{4.26} \\
          \multicolumn{1}{c}{} &\multicolumn{1}{c}{}&\multicolumn{1}{c}{}&\multicolumn{1}{c}{}&\multicolumn{1}{c}{}&\multicolumn{1}{c}{}&\multicolumn{1}{c}{}&\multicolumn{1}{c}{}&\multicolumn{1}{c}{}
          \\
    % \hline
    \multicolumn{1}{c}{\multirow{2}[2]{*}{w/o percep loss}} & \multirow{2}[2]{*}{2003462} & \multirow{2}[2]{*}{70.12} & \multirow{2}[2]{*}{78922} & \multirow{2}[2]{*}{272609} & \multirow{2}[2]{*}{7310412} & \multirow{2}[2]{*}{971504} & \multirow{2}[2]{*}{7027348} & \multirow{2}[2]{*}{4.31} \\
              \multicolumn{1}{c}{} &\multicolumn{1}{c}{}&\multicolumn{1}{c}{}&\multicolumn{1}{c}{}&\multicolumn{1}{c}{}&\multicolumn{1}{c}{}&\multicolumn{1}{c}{}&\multicolumn{1}{c}{}&\multicolumn{1}{c}{}
          \\
    \hline
    \multicolumn{1}{c}{\multirow{2}[2]{*}{\textbf{Ours}}} & \multirow{2}[2]{*}{\textbf{3121676}} & \multirow{2}[2]{*}{\textbf{102.28}} & \multirow{2}[2]{*}{\textbf{85607}} & \multirow{2}[2]{*}{\textbf{331483}} & \multirow{2}[2]{*}{\textbf{8990130}} & \multirow{2}[2]{*}{\textbf{1264451}} & \multirow{2}[2]{*}{\textbf{7667062}} & \multirow{2}[2]{*}{\textbf{4.75}} \\
            \multicolumn{1}{c}{} &\multicolumn{1}{c}{}&\multicolumn{1}{c}{}&\multicolumn{1}{c}{}&\multicolumn{1}{c}{}&\multicolumn{1}{c}{}&\multicolumn{1}{c}{}&\multicolumn{1}{c}{}&\multicolumn{1}{c}{}
          \\
    \hline
    \end{tabular}
    }%
  \label{tab2}%
\end{table*}%

\subsection{Experimental Results}

\subsubsection{Restoration results}
The results of teeth restoration are shown in Fig. \ref{exp1}. Our method displays the effectiveness of the restored teeth region in four Talking Face Generation methods, including Wav2lip \cite{PrajwalMNJ20}, MakeItTalk \cite{abs-2004-12992}, PC-AVS \cite{ZhouSWL0021}, and IP-LAP \cite{iplap}. It can be seen that both sharpness and details are obtained after restoring teeth. Wav2lip and MakeItTalk generate blurred areas of the teeth region. After applying our method, it is clear that our method has significant improvements in terms of sharpness, color accuracy, and level of detail in the resulting images. Although PC-AVS and LP-LAP methods are able to generate satisfactory teeth restoration results, our method further improves the clarity, texture, and color of the tooth region. In addition, our method preserves both frame coherence and lip-synchronization, which is friendly handling of every frame in talking face generation videos.

To further verify the teeth restoration robustness for arbitrary talking face generation methods, we evaluate our method in side face talking face videos and show the restoration results in Fig. \ref{exp_celian}. we observe that our model produces both sharpness and color teeth details.

\subsubsection{Quantitative results}
We compare our model with three recent state-of-the-art methods, including ESRGAN \cite{esrgan}, GFPGAN \cite{WangLZS21}, and Real-ESRGAN \cite{WangXDS21}. Table \ref{tab1} shows the quantitative results of our method and its competitors. Although ESRGAN obtains the highest variance while our method has a second performance, ours surpassed the state-of-the-art ones among other clarity metrics. With the input and output resolutions kept the same, our method achieves an impressive inference time of only 0.018 seconds per frame. This makes it more than three times faster than the fastest super-resolution restoration method available.

\subsubsection{Qualitative comparisons}
To present the superiority of our method, we provide generated samples compared with ESRGAN, GFPGAN, and Real-ESRGAN. As shown in Fig. \ref{exp_hq}, face restoration based on super-resolution is not well repaired in frame coherence and tooth texture details, our method has better performance visually in teeth texture details.

\subsection{Ablation Study}
In order to validate the effect of each component of our method, we conduct ablation study experiments for our HDTR-Net. Specifically, we set 2 conditions: (1) w/o CF: we replace channel fusion with convolution. (2) w/o branches below FGFF module: we remove the branches below FGFF in HDTR-Net. (3) w/o percep loss: we remove perceptual loss in the training stage. 

Table \ref{tab2} illustrates the qualitative results of ablation experiments. In our condition of Ours w/o reference FGFF module,  it is clear to see a significant reduction in results without the reference FGFF module. In our condition of Ours w/o perceptual loss, we also observe a decline in all of our evaluation metrics. 
We argue that adding CF and ref FGFF can obtain rich structural information and generate images with more high-frequency information. In addition, each component operates effectively in our method.

\section{Conclusion}
In this paper, we propose a real-time High-Definition Teeth Restoration Network (HDTR-Net), including two parallel Fine-Grained Feature Fusion modules and a Decoder module, to realize rich detail and texture in and around the teeth. The Fine-Grained Feature Fusion module is designed to merge low-level edge features and deep-level semantic features in feature dimensions to preserve the image refinement feature. In the inference stage, for two parallel FGFFs, the frame-by-frame guide input of the reference FGFF is capable of repairing teeth while ensuring frame coherence. The Decoder module is designed to merge the extracted feature maps from FGFF and restore the teeth region. With the combination of Fine-Grained Feature Fusion and Decoder, our method preserves more textural details and high-frequency information. Extensive qualitative and quantitative experiments have validated the performance of our method in arbitrary talking face generation methods without suffering lip synchronization and frame coherence. Compared to the super-resolution based face  restoration methods, our inference speed is three times faster or even higher.

\section*{Acknowledgements}
This work was supported by National Natural Science Foundation of China (No. 62172218), Shenzhen Science and Technology Program (No. JCYJ20220818103401003, No. JCYJ20220530172403007), Natural Science Foundation of Guangdong Province (No. 2022A1515010170).

\bibliographystyle{splncs04}
\bibliography{HDTR}
\end{document}